\documentclass[runningheads]{llncs}

\usepackage{amsmath,amssymb}
\usepackage{multirow}
\usepackage{enumerate}
\usepackage{makecell}
\usepackage{xcolor,lipsum,subcaption}
\usepackage[T1]{fontenc}
\usepackage{graphicx}

\begin{document}
\title{Sparsity- and Hybridity-Inspired \\ Visual Parameter-Efficient Fine-Tuning \\ for Medical Diagnosis}
%
%
\titlerunning{SH-PEFT for Medical Diagnosis} 
\author{Mingyuan Liu\inst{1} \and Lu Xu\inst{1} \and Shengnan Liu\inst{1} \and Jicong Zhang\inst{1,2,*}}
\authorrunning{M. Liu, L. Xu, S. Liu, and J. Zhang.}

\institute{$^{1}$School of Biological Science and Medical Engineering, \\ Beihang University, Beijing, China\\$^{2}$Hefei Innovation Research Institute, Beihang University, Hefei, Anhui, China
\email{$\{$liumingyuan95, xulu181221, liushengnan, jicongzhang$\}$@buaa.edu.cn}}
%
%
\maketitle              
\begin{abstract}
The success of Large Vision Models (LVMs) is accompanied by vast data volumes, which are prohibitively expensive in medical diagnosis.
To address this, recent efforts exploit Parameter-Efficient Fine-Tuning (PEFT), which trains a small number of weights while freezing the rest.
However, they typically assign trainable weights to the same positions in LVMs in a heuristic manner, regardless of task differences, making them suboptimal for professional applications like medical diagnosis.
To address this, we statistically reveal the nature of sparsity and hybridity during diagnostic-targeted fine-tuning, i.e., a small portion of key weights significantly impacts performance, and these key weights are hybrid, including both task-specific and task-agnostic parts.
Based on this, we propose a novel Sparsity- and Hybridity-inspired Parameter Efficient Fine-Tuning (SH-PEFT). 
It selects and trains a small portion of weights based on their importance, which is innovatively estimated by hybridizing both task-specific and task-agnostic strategies.
Validated on six medical datasets of different modalities, we demonstrate that SH-PEFT achieves state-of-the-art performance in transferring LVMs to medical diagnosis in terms of accuracy. 
By tuning around 0.01$\%$ number of weights, it outperforms full model fine-tuning. 
Moreover, SH-PEFT also achieves comparable performance to other models deliberately optimized for specific medical tasks.
Extensive experiments demonstrate the effectiveness of each design and reveal that large model transfer holds great potential in medical diagnosis.

\keywords{Parameter-efficient fine-tuning \and Medical diagnosis \and Vision transformer \and Sparsity and hybridity.}
\end{abstract}
\section{Introduction}

With the support of the vision transformer and massive data, large visual models (LVMs) have achieved great success~\cite{clip,vit}. 
However, when it comes to professional tasks such as medical diagnosis, the performance of LVMs is still insufficient.
Training medical-specific LVMs is prohibitively expensive, due to the difficulties of acquiring a large volume of medical data~\cite{trans-med,trans-repcnn,trans-mis}. 
To address this, Parameter-Efficient Fine-Tuning (PEFT) is proposed, which tunes only a small fraction of weights while freezing the rest~\cite{peft-fs,lora,peft-lm}.
It promotes effective knowledge transfer, reduces optimization difficulty, saves storage burden, and avoids over-fitting. 

\begin{figure}[t]
\includegraphics[width=\textwidth]{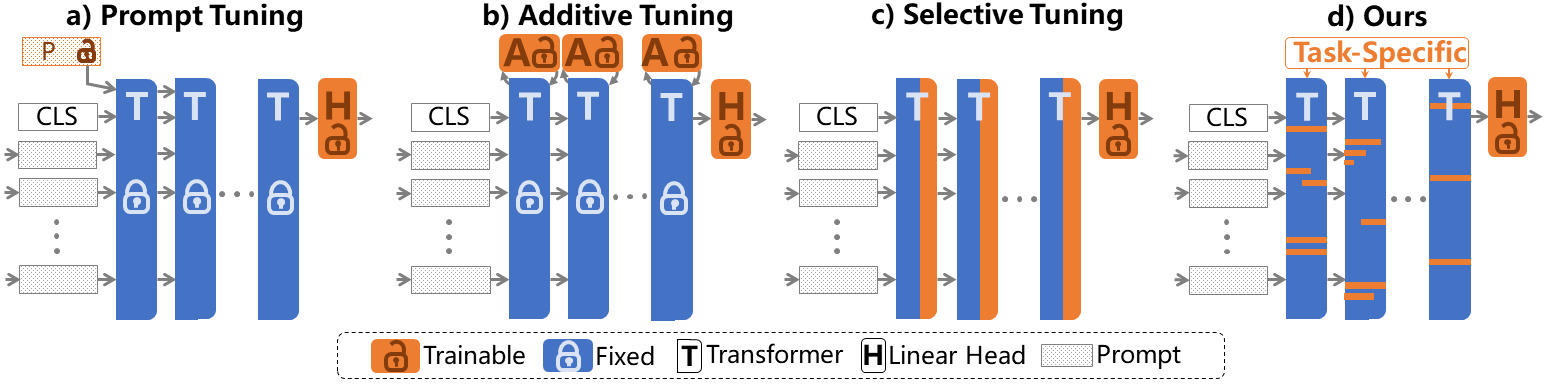}
\caption{
During PEFT, instead of heuristically assigning trainable weights to fixed positions across various medical diagnostic tasks, we employ a data-driven approach to select key weights for each task, enabling more effective fine-tuning.} 
\label{fig:intro}
\end{figure}

Recent PEFT efforts heuristically assign trainable weights to the same positions in LVMs. As shown in Fig.~\ref{fig:intro}, they could be roughly divided into three categories:  
\textbf{(1) Prompt Tuning} introduces trainable tokens while maintaining the pre-trained weights frozen~\cite{pt}. Techniques, such as the position of trainable tokens~\cite{vpt,ppt} and operations performed on them~\cite{de-pt}, are explored. Furthermore, the generalization capabilities are validated across different tasks, such as image generation~\cite{pt-gen}, image segmentation~\cite{pt-seg}, and video understanding~\cite{pt-video}.
\textbf{(2) Additive Tuning} inserts new trainable modules (a.k.a, adapters) either between or alongside existing transformer blocks~\cite{spt-adp}. A representative example is AdaptFormer~\cite{adapter}, which replaces all MLP blocks with encode-decoder modules. Moreover, unlike most designs that incur additional computational overhead during inference, LoRA~\cite{lora} utilizes low-rank decomposition to integrate adapters into the original LVM, avoiding extra computational cost.
\textbf{(3) Selective Tuning} trains a subset of weights within a LVM without changing the overall network structure, such as training all biases~\cite{bitfit,sel-bias}, attention layers~\cite{ft-att}, or normalization layers~\cite{ft-ln}, while keeping the remaining weights frozen.
\textbf{However}, the aforementioned methods add trainable parameters in a fixed manner to the same locations, ignoring the diverse downstream tasks and image modalities, therefore leading to suboptimal performance.

Motivated by this, we first conduct a statistical exploration of weight changes between pre-trained and fully fine-tuned LVMs on medical diagnosis. Our analysis reveals two significant findings: \textbf{(1) Sparsity} indicates that a minority of key weights play a majority role in downstream adaptation. \textbf{(2) Hybridity} means the positions of these key weights partially overlap across tasks, including both task-specific and task-agnostic components.
Based on the findings, we introduce a novel strategy called Sparsity- and Hybridity-inspired visual PEFT (SH-PEFT) for adapting LVMs to medical diagnosis. 
SH-PEFT selects a subset of weights that potentially introduce significant impacts on performance for tuning, based on their estimated importance.
The importance of each weight is innovatively approximated based on its hybrid contributions: its task-specific role in minimizing loss for a specific downstream medical task, as well as its task-agnostic significance within a LVM.
The estimation criterion is simple and effective, requiring only about 120 seconds for a dataset with 10k images, using a ViT-B/16 transformer, which is far less than the subsequent training time.

Our contributions are three folds: 
\textbf{(1)} Based on our statistical analysis, we reveal the sparsity and hybridity characteristics that exist in the process of transferring a large vision model to medical diagnosis.
\textbf{(2)} We propose a novel Sparsity- and
Hybridity-inspired Parameter Efficient Fine-Tuning (SH-PEFT), which estimates the importance of each weight by hybridizing both task-specific and task-agnostic strategies, and subsequently selects a small number of the most important weights for effective tuning. 
\textbf{(3)} Extensive experiments on six medical datasets reveal the effectiveness of our SH-PEFT: it achieves state-of-the-art PEFT performance under a comparable number of trainable weights; it performs comparably with models deliberately designed for specific diagnostic tasks; its hybrid weight importance estimation strategy effectively enhances performance. 

\begin{figure}[t]
\includegraphics[width=\textwidth]{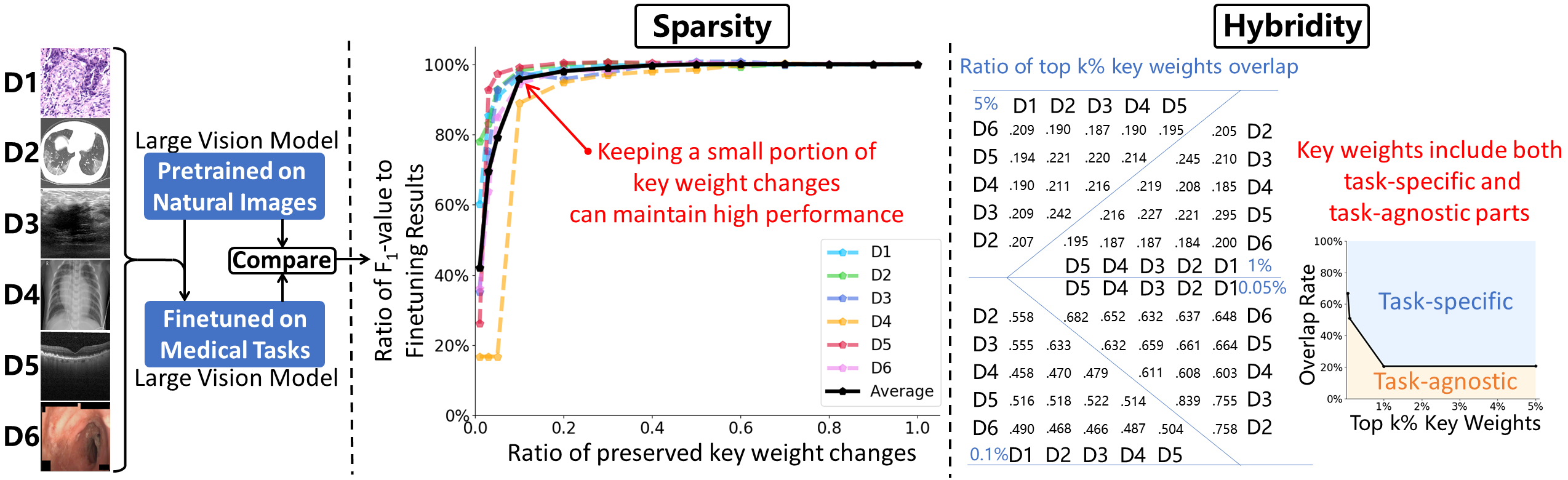}
\caption{
We experimentally conclude the sparsity and hybridity nature of key weight distributions from six medical datasets, by comparing differences between pre-trained and medical fine-tuned CLIP model.
\textbf{Sparsity} indicates that a few key weights largely impact performance, thereby motivating us to select important weights for tuning.
\textbf{Hybridity} indicates that key weights contain both task-specific and task-agnostic parts, thereby prompting us to explore a hybrid strategy to locate key weights for more effective PEFT.} 
\label{fig:sta}
\end{figure}

\section{Method}
\subsection{Statistical Evidence of Sparsity and Hybridity}
For constructing effective PEFT for medical diagnosis, we first explore the feasibility of tuning a few key weights and the proper location to introduce these trainable weights.
As shown in Fig.~\ref{fig:sta}, we experimentally reveal the existence of sparsity and hybridity nature of the key weight distributions, by analyzing weight differences between pre-trained and medical fine-tuned models. 

Specifically, experiments are conducted on six medical datasets with different modalities, 
to maximize the applicability of our conclusions to various diagnostic tasks.
Examples of six datasets are shown in Fig.~\ref{fig:sta}. 
We choose CLIP~\cite{clip} as the LVM due to its widespread application and outstanding performance among publicly available checkpoints.
More details are elaborated in Sec.~\ref{sec:detail}.

Sparsity indicates that a small ratio of key weights significantly impacts the performance of knowledge transfer, suggesting the potential for selective fine-tuning in medical PEFT. 
Given pre-trained weights $\mathcal{W}^{ori}$ of a model $\Phi_{\mathcal{W}^{ori}}(\cdot)$ and fine-tuned weights $\mathcal{W}^{ft}$ on a medical task, $\Delta\mathcal{W}^{ft}=|\mathcal{W}^{ft}-\mathcal{W}^{ori}|$ measures weight changes.
We first identify top $k\%$ elements with the largest variations and directly replace weights in $\mathcal{W}^{ori}$ at positions of top $k\%$ by the corresponding weights from $\mathcal{W}^{ft}$, denoted as $\mathcal{W}^{ori\gets ft@ k\%}$. 
Then performance of $\Phi_{\mathcal{W}^{ori\gets ft@ k\%}}(\cdot)$ is directly validated, and results are shown in Fig.~\ref{fig:sta}. 
Results show that keeping around 10$\%$ of the changes could maintain around 95$\%$ of the full fine-tune performance, revealing the feasibility of selective tuning for medical PEFT.

Hybridity indicates that key weights contain both task-specific and task-agnostic parts, suggesting the necessity of proposing a hybrid strategy for locating key weights. Given fine-tuned weights on different medical diagnostic tasks $\mathcal{W}^{ft}_{T_m}$ and $\mathcal{W}^{ft}_{T_n}$, we measure the positional overlap of the key weights between them at different $k\%$ ($5\%$, $1\%$, $0.1\%$, and $0.05\%$). 
Results in Fig.~\ref{fig:sta} show that, in different tasks, most of the key weights do not overlap, indicating they are task-specific. Meanwhile, positions of a small portion of key weights are shared across different tasks, suggesting they are task-agnostic. 
This inspires us to hybrid both task-specific and task-agnostic strategies to identify the key weights.

\subsection{Hybrid Weight Importance Estimation for PEFT}

\begin{figure}[t]
\includegraphics[width=\textwidth]{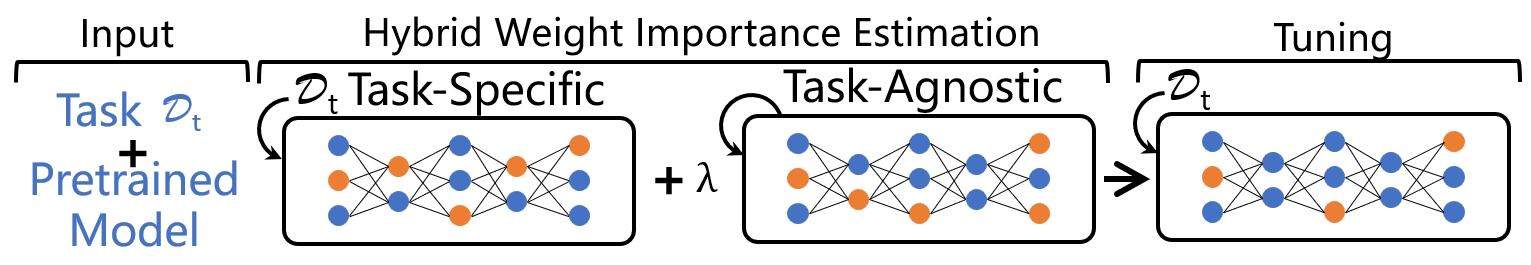}
\caption{
Inspired by the sparsity and hybridity, we propose a novel SH-PEFT approach to fine-tune a few key weights for adapting pre-trained vision transformers to medical diagnosis. 
The key weights can be effectively and quickly identified by jointly considering their importance from both task-specific and task-agnostic perspectives.
} 
\label{fig:model}
\end{figure}

Inspired by the aforementioned findings, we propose SH-PEFT to adaptively determine trainable weights in a model, by jointly considering their importance in both  the specific task and the model structure, as shown in Fig.~\ref{fig:model}

Given a dataset $\mathcal{D}_t$, the learning objective is to minimize the empirical risk $E(\mathcal{D}_t, \mathcal{W})$ by updating weights $\mathcal{W}$ in a model.
For a model with $m$ layers, $\mathcal{W}=\{ \mathbf{w}_1, \mathbf{w}_2, ..., \mathbf{w}_m\}$ and 
the $n$-th weight in layer $m$ is $\textbf{w}_{m,n}$. Its importance $\mathbb{I}_{m,n}$ could be estimated by:
\begin{equation}
\begin{aligned}
\mathbb{I}_{m,n} &= \mathbb{I}^{td}_{m,n}+\lambda\mathbb{I}^{ta}_{m,n}  \\
&= | \Delta E(\mathcal{D}_t, \mathcal{W}| \textbf{w}_{m,n}\to\hat{\textbf{w}}_{m,n})| + \lambda|\Delta E(\mathcal{D}, \mathcal{W}| \textbf{w}_{m,n}\to0)|
\end{aligned}
\end{equation}

The first term $\mathbb{I}^{td}_{m,n}$ is task-dependent. It measures the change of empirical risk caused by the weight update from  $\textbf{w}_{m,n}$ to $\hat{\textbf{w}}_{m,n}$ after training on $\mathcal{D}_t$. 
The second term $\mathbb{I}^{ta}_{m,n}$ is task-agnostic. It measures the empirical risk change by removing a weight $\textbf{w}_{m,n}$ on arbitrary task $\mathcal{D}$, reflecting the significance of a weight $\textbf{w}_{m,n}$ in the model. $\lambda$ balances the values between the two terms. 
However, the two terms are difficult to estimate. 
Because they require training or evaluating a model several times for each weight, which is computationally prohibitive. 

To address this issue, we estimate the empirical risk changes in two ways.
The first way is intuitive. We use the accumulation of gradients over several iterations to judge the trend of weight changes, i.e., $\mathbb{I}^{td-L1}_{m,n} = \Sigma_{b=1}^{B} \partial E_b / \partial \textbf{w}_{m,n}$, where $B$ is the number of iteration and $\partial E_b / \partial \textbf{w}_{m,n}$ calculates the gradient on each mini-batch of weight $\textbf{w}_{m,n}$. $L1$ means it is the first estimation method.
The task-agnostic part is estimated by $\mathbb{I}^{ta-L1}_{m,n}=|\textbf{w}_{m,n}|$, which is the absolute value of the weight. 
It is based on the assumption that a large weight can cause significant changes to the input features so that it plays a more significant role in maintaining the functionality of the model than a small weight. This assumption is previously applied in the model pruning tasks~\cite{prun-imp,prun-platon} and is transferred to PEFT scenario by our SH-PEFT. 

The second way is inspired by~\cite{spt}, where the task-dependent importance of $\textbf{w}_{m,n}$ could be estimated by its first-order Taylor expansion in its vicinity range. It could be written as $\mathbb{I}^{td-L2}_{m,n}=\partial \mathcal{L} / \partial \textbf{w}_{m,n}*(\hat{\textbf{w}}_{m,n}-\textbf{w}_{m,n})$. Since the estimation should be finished within a few forward passes, the weight differences could be abbreviated as its gradient, so that $\mathbb{I}^{td-L2}_{m,n}\approx(\partial \mathcal{L} / \partial \textbf{w}_{m,n})^2$. 
To avoid inconsistent scaling between the two terms due to squaring, we also apply squaring to the task-agnostic part $\mathbb{I}^{ta-L2}_{m,n}=(\textbf{w}_{m,n})^2$. 
Moreover, in both ways, before applying $\lambda$, the second term is weighted by $\Sigma\mathbb{I}^{td}_{m,n}/\Sigma\mathbb{I}^{td}_{m,n}$ to balance their scaling differences.

After estimating the importance of all parameters, we set a threshold $\tau$ based on the number of trainable weights to be allocated. If the importance of a weight is larger than $\tau$, the weight can be updated; if it is smaller, it remains unchanged.
Therefore, the update strategy for each weight at step t+1 is:
\begin{equation}
\textbf{w}_{m,n}^{t+1} = \textbf{w}_{m,n}^{t} - \eta \frac{\partial \mathcal{L}}{\partial \textbf{w}_{m,n}^{t}}\mathbf{M}_{m,n}~,where~\mathbf{M}_{m,n}= \left\{
\begin{array}{rcl}
1       &      & if~\mathbb{I}_{m,n}>\tau\\
0     &      & else
\end{array} \right.
\end{equation}
, where $\mathbf{M}_{m,n}$ is a binary mask of weight $\textbf{w}_{m,n}$ and $\eta$ is the learning rate.

It is worth noting that, although we gain inspiration from previous works~\cite{spt,prun-imp,prun-platon}, we have unique contributions. 
Specifically, our SH-PEFT differs from the most similar work, SPT~\cite{spt}, in three aspects:
\textbf{1) The scope of weight selection}: SPT selects weights only from linear layers for data-specific tuning. However, other layers like layer-norm also play an important role in PEFT~\cite{ssf,ft-ln}. Our SH-PEFT extends the selection scope to all operations, which enhances the model adaptability to downstream tasks. 
\textbf{2) The strategy of weight importance estimation}: SPT uses the square of the derivative to estimate the weight importance in a task-specific manner. In contrast, we draw inspiration from our discovered hybridity nature and jointly consider both task-specific and -agnostic factors, leading to more reliable weight estimation strategy for better PEFT.
\textbf{3) The mode of usage}: SPT works in conjunction with other PEFT methods like adapters~\cite{spt-adp}. Differently, our SH-PEFT could work independently, which flexibly circumvents the shortage of other methods, such as the computational overhead introduced by an adapter.

\section{Experiment and Result}
\subsection{Training Details}\label{sec:detail}
Six datasets with different modalities are used for our statistical analysis and quantitative evaluation in order to ensure the applicability of the conclusion and method to diverse medical diagnostic problems. They include:
\textbf{D1}: Chaoyang~\cite{chaoyang} is a \textit{pathological} dataset of the human colon. It has 4,021 training and 2,139 testing images, covering 4 categories: normal, serrated, adenocarcinoma, and adenoma. 
\textbf{D2}: Covid19-CT~\cite{covid19-ct} is lung \textit{CT} dataset for diagnosing Covid-19. It has 425/118/203 images for training/validation/testing, respectively. 
\textbf{D3}: BUSI~\cite{busi} is an \textit{ultrasound} image dataset for early diagnosis of normal, benign, or malignant breast cancer, with 559/79/160 images for training/validation/testing.
\textbf{D4}: CXT3~\cite{oct} is a chest \textit{X-ray} dataset from children including normal, bacterial, and viral cases, with 4,708/524/1,248 images for training/validation/testing. 
\textbf{D5}: OCT~\cite{oct} is a retina \textit{OCT} dataset, with 97,477/10,832/1,000 images for training/validation/testing. including choroidal neovascularization, diabetic
macular edema, multiple drusen, and normal.
\textbf{D6}: LIUMC~\cite{liumc} is a \textit{colonoscopy} dataset for ulcerative colitis with 4 securities. It has 9,590 images for training and 1,686 images for testing.

Experiments are conducted on CLIP~\cite{clip} pre-trained visual transformer of ViT-B/16 structure. This is because its innovative training approach, which connects images and text, has become a paradigm for other large-model training, as well as due to its excellent performance among publicly available checkpoints. During training, the final projection layer in vision transformer is replaced by a $L_2$ normalization and a linear layer.
The training uses the SGD optimizer with batch size 64. 
The initial learning rate is 0.001 and is adjusted by CosineAnnealing. 
Each dataset is trained for 40k iterations, and $F_1$-value measures the final performance. The key weights are selected within one training epoch.
Unless otherwise specified, the SH-PEFT model uses $\mathbb{I}^{L2}$ strategy with $\lambda=1$, and selects 1$\%$ of the parameters as trainable parameters in the following experiments

\begin{table}[t]
    \renewcommand{\arraystretch}{1.3}
    \caption{Comparison with state-of-the-art PEFT methods, measured by $F_1$ ($\%$). `S', `A', and `P', denote selective, additive, and prompt tuning respectively.}
    \label{tab:sota}
    \centering
    \begin{tabular}{c|c|p{0.8cm}<{\centering} p{0.8cm}<{\centering} p{0.8cm}<{\centering} p{0.8cm}<{\centering} p{0.8cm}<{\centering} p{0.8cm}<{\centering} | p{0.8cm}<{\centering}}
    \hline
    Method (Pub'Year) & Type & $\mathcal{D}$1 & $\mathcal{D}$2 & $\mathcal{D}$3 & $\mathcal{D}$4 & $\mathcal{D}$5 & $\mathcal{D}$6 & Avg  \\
    \hline
    Full Finetune & S & 80.2 & 73.5 & 82.5 & 76.3 & 94.2 & 69.1 & 79.3 \\
    Linear Prob & S & 68.3 & 76.4 & 76.0 & 79.6 & 81.9 & 60.4 & 73.8 \\
    Adapter-par (NeurIPS'22)~\cite{adapter} & A & 75.5 & 78.6 & 84.1 & 77.7 & 93.2 & 67.4 & 79.4 \\ 
    SSF (NeurIPS'22)~\cite{ssf} & A & 78.8 & 79.5 & 89.8 & 78.5 & 92.2 & 71.9 & 81.8 \\
    LoRa (ICLR'22)~\cite{lora} & A & \textbf{81.9} & 82.4 & 87.8 & 78.0 & \textbf{96.3} & 67.9 & 82.4 \\
    VPT-Deep (ECCV'22)~\cite{vpt} & P & 70.4 & 78.2 & 75.1 & 78.0 & 82.3 & 64.2 & 74.7 \\ 
    VPT-Shallow (ECCV'22)~\cite{vpt} & P & 74.2 & 78.8 & 81.6 & 80.3 & 90.0 & 69.3 & 79.0 \\
    FT-LN (Arxiv'23)~\cite{ft-ln} & S & 74.2 & 75.5 & 79.6 & 76.4 & 87.2 & 68.1 & 76.8 \\ 
    BitFit (ACL'22)~\cite{bitfit} & S & 79.1 & 80.0 & 86.1 & 73.6 & 93.3 & 72.2 & 80.7 \\
    FT-Att (ECCV'22)~\cite{ft-att} & S & 81.1 & 81.6 & 88.8 & 79.4 & 95.8 & 70.1 & 82.8 \\ 
    SPT-LoRa (ICCV'23)~\cite{spt} & S+A & 81.8 & 80.2 & 90.2 & 77.3 & 96.0 & 69.5 & 82.9 \\
    \hline
    SH-PEFT (Ours) & S & 80.6 & \textbf{83.0} & \textbf{90.5} & \textbf{81.0} & 95.6 & \textbf{72.7} & \textbf{83.9} \\
    \hline
    \end{tabular}
\end{table}

\begin{table}[t]
\begin{minipage}{.50\textwidth}
  \centering
  \captionsetup{justification=centering}
  \caption{Comparison with latest \\ efforts on Chaoyang dataset.}
  \label{tab:chaoyang}
    \begin{tabular}[t]{c|c c}
    \hline
    Method & $F_1\%$ & ACC$\%$ \\
    \hline
    NSHE(TMI'22)~\cite{chaoyang} & 76.5 & 83.4 \\
    PVB+L(ECCV'22)~\cite{chaoyang-pvb} & - & 84.3 \\
    GSB(NN'24)~\cite{chaoyang-gsb} & - & 82.5 \\
    \hline
    SH-PEFT (Ours) & \textbf{80.6} & \textbf{84.8} \\
    \hline
    \end{tabular}
\end{minipage}%
\begin{minipage}{.50\textwidth}
  \centering
  \captionsetup{justification=centering}
  \caption{Comparison with latest \\ efforts on COVID19-CT dataset.}
  \label{tab:covid19}
  \begin{tabular}[t]{c|c c}
    \hline
    Method & $F_1\%$ & ACC$\%$ \\
    \hline
    SKNet(CVPR'19)~\cite{covid-sknet} & 76 & 77 \\
    ECAN (ECCV'20)~\cite{covid-eca} & 74 & 75 \\
    ResGANet(MIA'22)~\cite{covid-resganet} & 81 & 80 \\
    \hline
    SH-PEFT (Ours) & \textbf{83.0} & \textbf{83.3} \\
    \hline
    \end{tabular}
\end{minipage}
\end{table}

\subsection{Comparison with State-of-the-art Methods}
\textbf{Superiority over PEFT methods}: 
Table~\ref{tab:sota} shows our SH-PEFT outperforms PEFT methods of different types on six medical datasets measured by $F_1$-value (Due to space limitations, ACC and AUC results are shown in Supp.).
It demonstrates that our method can effectively transfer general visual knowledge from LVMs to medical diagnosis, indicating the effectiveness of our flexible selective learning method and hybrid feature weight importance estimation strategy.
Specifically, for fair comparisons, all models are trained following their official implementation and use the same hyper-parameters mentioned in Sec.~\ref{sec:detail}. 

\noindent\textbf{Superiority over Domain-specific Methods}: 
Table~\ref{tab:chaoyang} and Table~\ref{tab:covid19} show that our SH-PEFT achieves comparable results to recent deep learning methods that are optimized for these specific tasks (Due to space limitations, more results of other datasets are shown in Supp.). 
The outstanding outcomes indicate that our SH-PEFT is effective and
designing medical-targeted PEFT could be a promising approach for better medical diagnostic applications.

\begin{figure}[t]
\centering
    \includegraphics[width=0.65\textwidth]{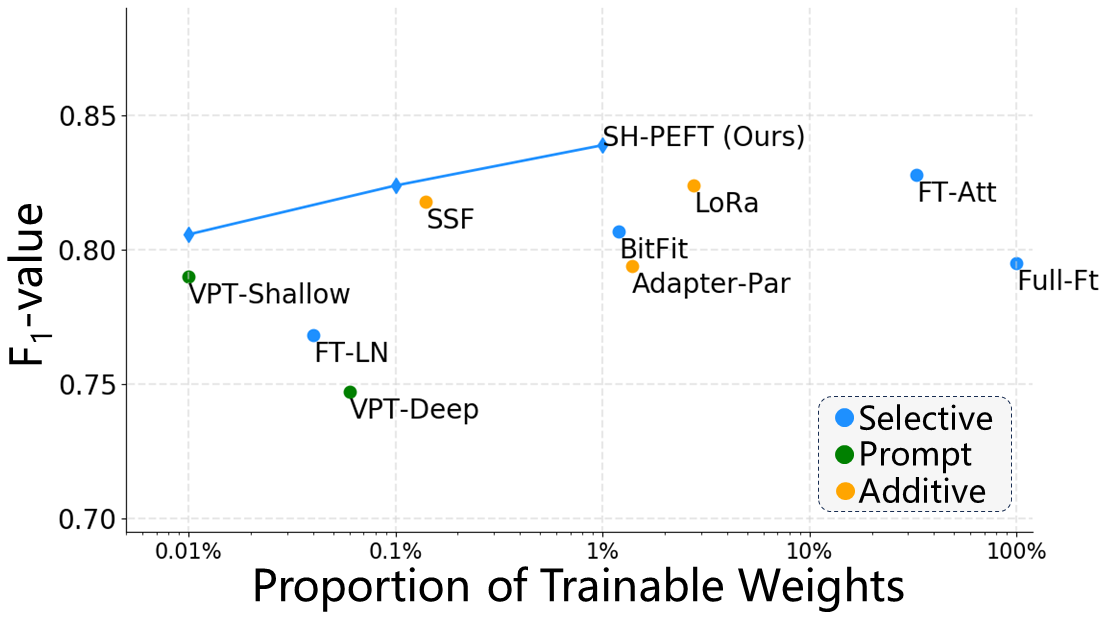}
    \caption{Under the same ratio of trainable weights, SH-PEFT outperforms state-of-the-art PEFT methods in terms of the average $F_1$-value across six datasets.}
    \label{fig:result}
\end{figure}

\noindent\textbf{Superiority Under Comparable Number of Trainable Weights}: Fig.~\ref{fig:result} demonstrates that, compared to other methods, our SH-PEFT achieves better performance when the number of trainable parameters is similar. 
It is worth noting that tuning around 0.01$\%$ number of weights by SH-PEFT outperforms full model fine-tuning, indicating that our allocation of trainable parameters is effective.

\begin{table}[t]
\begin{minipage}{.49\textwidth}
    \centering
    \captionsetup{justification=centering}
    \caption{Ablation on hybrid estimation strategies, measured on the average of six datasets.}
    \label{tab:hybrid}
    \begin{tabular}{c|c|c c}
    \hline
    $\mathbb{I}^{ta}$ & $\mathbb{I}^{td}$ & $F_1\%$ & ACC$\%$ \\
    \hline
    $\checkmark$ &  & 81.9 & 84.5 \\
    & L1 & 81.3 & 84.3 \\
    $\checkmark$ & L1 & 82.4 & 85.0 \\
    & L2 & 82.0 & 84.7 \\
    $\checkmark$ & L2 & \textbf{83.9} & \textbf{86.3} \\
    \hline
    \end{tabular}
\end{minipage}
\begin{minipage}{.49\textwidth}
    \centering
    \captionsetup{justification=centering}
    \caption{Ablation on the balancing weight $\lambda$, measured on the average of six datasets.}
    \label{tab:lambda}
    \begin{tabular}{c|c c}
    \hline
    $\lambda$ & $F_1\%$ & ACC$\%$ \\
    \hline
    1.5 & 82.4 & 85.1 \\
    1.2 & 83.1 & 85.6 \\
    1.0 & \textbf{83.9} & \textbf{86.3} \\
    0.8 & 82.6 & 85.1 \\
    0.5 & 82.5 & 85.1 \\
    \hline
    \end{tabular}
\end{minipage}
\end{table}

\subsection{Ablation Studies}
\textbf{Hybrid weight importance estimation}: Table~\ref{tab:hybrid} demonstrates that task-specific and task-agnostic strategies could work complementarily to improve performance, indicating the effectiveness of our hybrid strategy.

\noindent\textbf{Effectivevness of $\lambda$}: Table~\ref{tab:lambda} demonstrates the performance is relatively robust to the selection of $\lambda$. 

\section{Conclusion}
We statistically reveal the characteristics of sparsity and hybridity when transferring general LVMs to medical diagnosis. Inspired by this, we propose SH-PEFT to allocate a small portion of trainable weights for tuning, based on their hybrid importance measured in both task-specific and task-agnostic manner.
We validate the effectiveness of PEFT on six medical diagnosis datasets with different modalities.
Results show that, with the same number of trainable weights, our SH-PEFT outperforms existing PEFT methods in terms of accuracy.
Furthermore, the model fine-tuned by SH-PEFT outperforms deep learning models specifically optimized for diagnostic tasks, indicating the effectiveness of our strategy.
Ablation studies further demonstrate the effectiveness of each design.


\bibliographystyle{splncs04}
\bibliography{refbib}

\end{document}